\documentclass[11pt]{article}
\usepackage[final]{acl}
\usepackage{times}
\usepackage{latexsym}
\usepackage[T1]{fontenc}
\usepackage[utf8]{inputenc}
\usepackage{microtype}
\usepackage{graphicx}
\usepackage{booktabs}
\usepackage{amsmath,amssymb}
\usepackage{array}
\usepackage{tabularx}
\usepackage{multirow}
\usepackage{enumitem}

\newcommand{\emo}{\textsc{Emo-Jev}}

\title{\emo: Probabilistic Reasoning for Emotion Classification with Jev}
\author{Yazhou Zhang \and Junhao Yu \\
  School of Computer Science and Technology, Tianjin University \\
  Tianjin, China}

\begin{document}
\maketitle
\begin{abstract}
Jev offers an alternative interface for language understanding:
given an input and predefined questions, it returns probabilistic
decisions rather than free-form responses. Whether this interface
can support effective reasoning for text classification against leading LLMs remains an
open questions. 
We introduce Emo-Jev, a training-free
framework with two complementary implementations.
Emo-Jev-D decomposes classification into task-specific atomic
judgments and composes their probabilities into a final prediction.
Emo-Jev-SC constructs multiple judgment paths from complementary
perspectives and aggregates their predictions into a consensus
decision. 
We evaluate Emo-Jev on eight datasets spanning sentiment analysis,
emotion recognition, sarcasm detection and humor
detection, comparing against direct Jev classification and five SoTA LLMs under input/output and chain-of-thought reasoning.
Standard Jev achieves 62.93\% average macro-F1 versus
67.28\% for the strongest LLM baseline, with lower
observed latency and generally lower cost.
\end{abstract}

\section{Introduction}
The advancement of large language models (LLMs) has transformed text classification into a generation task: given an input and candidate labels, an LLM produces a classification result by generating text~\citep{wang2026adaptive}. Reasoning-based approaches extend this process with intermediate analysis before the final prediction
\citep{wei2022chain,guo2026textreasoning}. However, generating reasoning traces introduces additional latency and cost. When the output space is predefined, a natural question
is whether effective reasoning can instead operate through probabilistic decisions?

\begin{figure}[t]
    \centering
    \includegraphics[width=\linewidth]{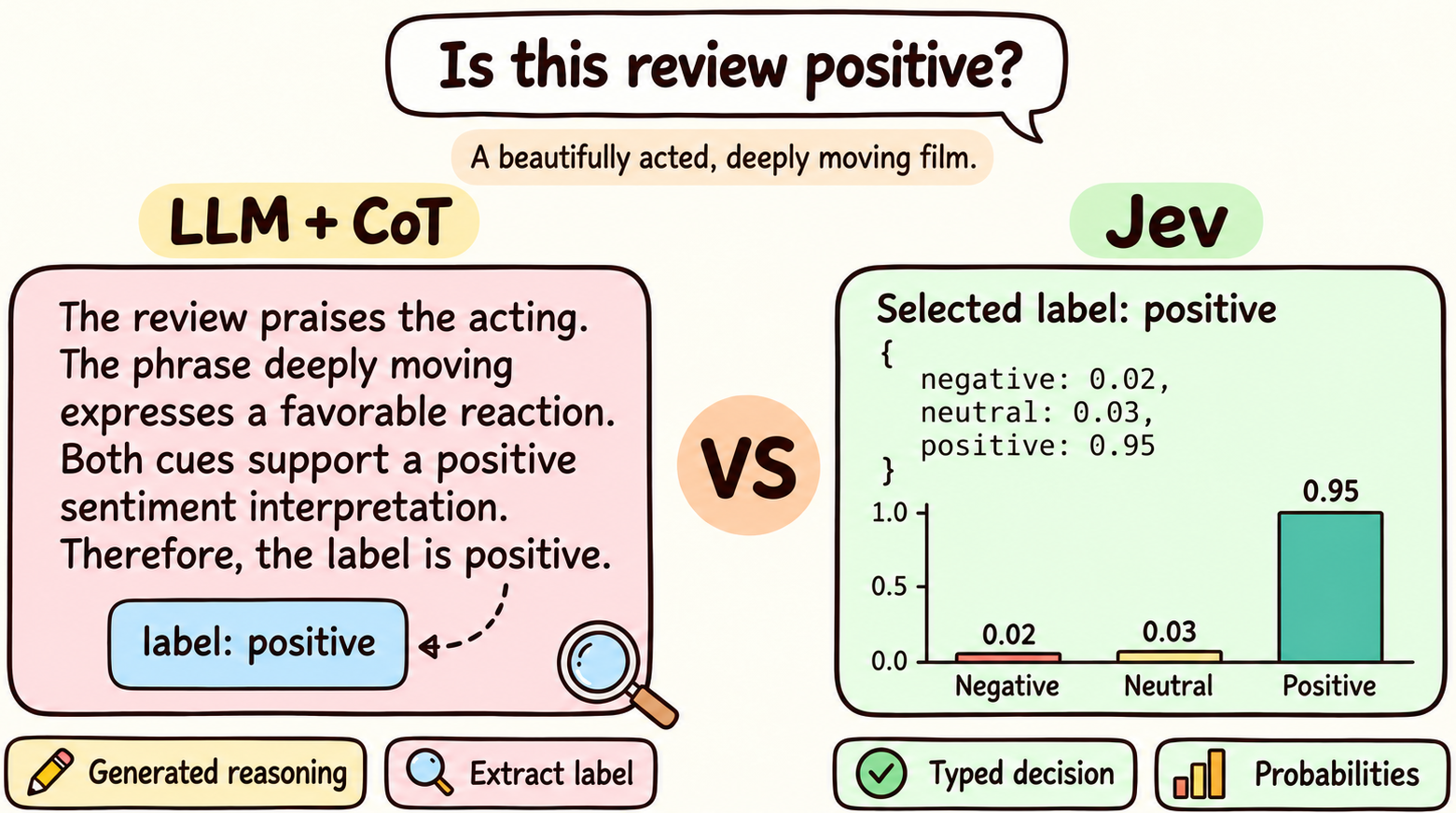}
    \caption{Text classification with an LLM and Jev.}
    \label{fig:example}
\end{figure}

Jev provides an interface for investigating this question, as shown in Fig.~\ref{fig:example}.
Given an input state and predefined questions, it returns typed decisions with associated probabilities rather than free-form responses \citep{almeida2026jev}.
This interface supports direct label prediction, but also exposes intermediate judgments that an external procedure can compose or aggregate. 
It therefore offers an opportunity to study reasoning over explicit decision probabilities. 

Two empirical questions motivate our study.
First, how does direct Jev classification compare with leading
language and reasoning models in predictive performance and
inference efficiency?
Second, can reasoning over Jev's probabilistic judgments improve
upon its direct predictions?

To address these issues, we introduce \textbf{Emo-Jev}, a training-free framework with
two implementations.
\textbf{Emo-Jev-D} decomposes classification into task-specific
atomic judgments and combines their probabilities through
predefined decision rules.
The judgments distinguish polarity from intensity in sentiment
analysis, assess individual emotions in emotion recognition,
examine literal-intended discrepancies in irony and sarcasm
detection, and identify cues such as incongruity and wordplay
in humor detection.
\textbf{Emo-Jev-SC} constructs multiple judgment paths from
complementary perspectives and aggregates their predictions
into a consensus decision.

We evaluate direct Jev classification and both Emo-Jev variants
on eight benchmarks covering sentiment analysis, emotion
recognition, sarcasm detection and humor detection.
Comparisons with five SoTA LLMs (i.e., GPT-6 Luna, Gemini 3.8 Flash, Qwen 3.7 Plus, DeepSeek V4.1 Flash, Doubao-Seed-2.0-mini) under input/output and chain-of-thought
prompting assess classification performance.
Standard Jev achieves 62.93\% average macro-F1,
compared with 67.28\% for the strongest generative
configuration, Gemini under IO, at a
96.41\% lower inference cost.
It also records the lowest median latency of 250 ms,
although it does not consistently reduce token usage.
Emo-Jev-D and Emo-Jev-SC achieve average macro-F1
scores of 62.00\% and 61.61\%, respectively,
with higher inference costs.
These results identify Standard Jev as an economical
classification alternative, while showing that
decomposition and multi-path aggregation do not
consistently improve its predictions.
The main contributions of this work are as follows:
\begin{itemize}
    \item We present the first systematic study of Jev for
    sentiment, emotion, sarcasm, and humor classification.
    \item We propose Emo-Jev-D and Emo-Jev-SC for decomposed
    and multi-path probabilistic reasoning.
\end{itemize}

\section{Related Work}
\subsection{LLM-Based Text Classification}
Research on LLM-based text classification broadly follows two
directions: task-specific model adaptation and inference-time
reasoning. The first incorporates task knowledge through
fine-tuning, knowledge transfer, and ensemble learning.
\citet{wang2026adaptive} proposed an adaptive boosting framework
that iteratively fine-tunes LLM-based classifiers and aggregates
their predictions. For conversational emotion recognition,
\citet{zhang2025dialoguellm} incorporated dialogue context and
emotion knowledge into model fine-tuning.
\citet{shvets2025emopillars} used LLM-generated emotion examples
to train compact classifiers.
\citet{lai2024rvisa} combined rationale generation and
verification to train a reasoner for implicit sentiment analysis.
These approaches incorporate task-specific supervision into
model parameters, using annotated data or generated examples.

The second direction improves classification through intermediate
reasoning at inference time.
\citet{sun2023carp} introduced Clue and Reasoning Prompting,
which identifies textual clues before inferring a label.
\citet{fei2023thor} decomposed implicit sentiment analysis
into successive judgments about aspects, opinions, and
sentiment polarity. For sarcasm detection,
\citet{yao2025sarcasmcue} investigated sequential and
non-sequential reasoning over linguistic, contextual, and
emotional cues, while \citet{lee2025pragmatic} incorporated
pragmatic interpretation and reflection.
Beyond individual reasoning paths,
\citet{wang2023selfconsistency} proposed aggregating answers
from multiple sampled reasoning paths.
Emo-Jev instead operates on explicit decision probabilities:
Emo-Jev-D composes task-specific atomic judgments, whereas
Emo-Jev-SC aggregates predictions from complementary judgment
paths. Both use a frozen Jev model without generating textual
reasoning traces.

\subsection{Jev for Structured Decisions}
Jev answers predefined questions about an input state with
typed decisions and associated probabilities
\citep{almeida2026jev}. Its initial provider-reported evaluations
cover security incident triage, agent trace review, invoice
processing, and customer service, using model-generated
reference judgments \citep{typesafe2026workflows}.
In computational social science,
\citet{ibrahim2026decision} evaluated decision models on
text annotation tasks against benchmark labels.
They found that Jev substantially reduced inference cost
but achieved lower macro-F1 than the strongest LLM on most
evaluation tasks.

Research on multi-stage decision procedures examines how
Jev's outputs support subsequent computation and model selection.
\citet{deng2026scientific} evaluated semantic relation selection
with Jev and its effects on downstream scientific calculations.
\citet{li2026judge} investigated Jev as a judge of model outputs
and introduced a cascade that delegates uncertain judgments
to a stronger model.
Our work investigates a different use of the decision interface:
composing atomic judgment probabilities and aggregating
alternative judgment paths for text classification.

\begin{figure*}[t]
    \centering
    \includegraphics[width=0.95\linewidth]{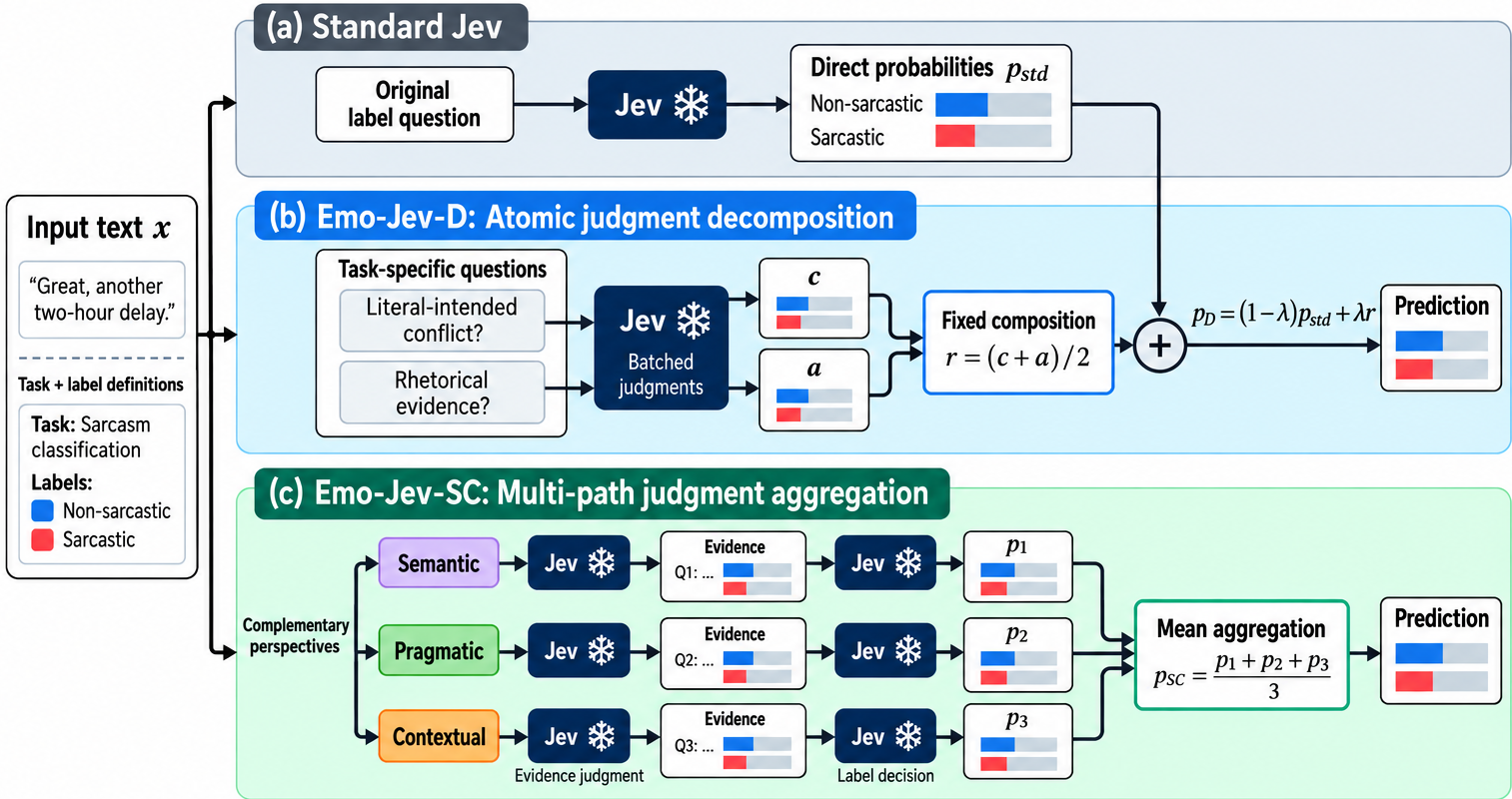}
    \caption{Overview of the proposed Emo-Jev.}
    \label{fig:framework}
\end{figure*}

\section{The Proposed Approach}
\subsection{Problem Definition}
Given a text $x$ and a predefined label set
$\mathcal{Y}=\{y_1,\ldots,y_K\}$, single-label classification
predicts
\begin{equation}
    \hat y=\arg\max_{y_k\in\mathcal{Y}}p_k(x)
\end{equation}
where $p_k(x)$ is the predicted probability of label $y_k$.
For multi-label classification, the prediction is a subset
$\hat{\mathcal{Y}}=\{y_k:p_k(x)\geq\tau_k\}$.
We consider sentiment analysis, emotion recognition,
irony and sarcasm detection, and humor detection.

\subsection{Framework Overview}

Emo-Jev performs classification through probabilistic judgments
obtained from a frozen Jev model.
Let $J_{\theta}(s,q)$ denote its response to a typed question
$q$ about an input state $s$.
A categorical question returns a distribution over predefined
options, whereas a binary question returns a probability
in $[0,1]$.
The parameters $\theta$ remain unchanged throughout inference.

We instantiate the framework in two ways.
\textbf{Emo-Jev-D} decomposes a task into atomic judgments
and maps their probabilities to label scores through
predefined rules.
\textbf{Emo-Jev-SC} evaluates the input through complementary
judgment paths and aggregates their label predictions.
The former specifies how intermediate judgments are composed,
whereas the latter combines alternative interpretations
of the same input.

\paragraph{Standard Jev.}
For single-label tasks, the direct question
$q_t^{\mathrm{std}}$ specifies the task, candidate labels,
and label definitions:
\begin{equation}
    \mathbf{p}^{\mathrm{std}}
    =J_{\theta}(x,q_t^{\mathrm{std}})
    \in\Delta^{K-1}
\end{equation}
where $\Delta^{K-1}$ is the probability simplex.
For multi-label tasks, we ask one binary question per label,
yielding $\mathbf{p}^{\mathrm{std}}\in[0,1]^K$.
These predictions constitute the Jev-Direct baseline.

\subsection{Emo-Jev-D: Atomic Judgment Decomposition}
An atomic question addresses a specific property relevant
to the classification decision.
For task $t$, we define a question inventory
$\mathcal{Q}_t=\{q_{t,1},\ldots,q_{t,m_t}\}$ and collect
\begin{equation}
    \mathbf{z}_t
    =\bigl(J_{\theta}(x,q_{t,1}),\ldots,
           J_{\theta}(x,q_{t,m_t})\bigr)
\end{equation}

A deterministic mapping converts these outputs into
task-specific label scores:
\begin{equation}
    \mathbf{r}_t=G_t(\mathbf{z}_t)
\end{equation}

\paragraph{Sentiment analysis.}
We separate the presence of an evaluative attitude,
its polarity, and its intensity.
Let $u$ denote the probability that the text expresses
a nonneutral evaluation of the target.
A second question assesses positive versus negative polarity
under the assumption that a nonneutral evaluation is present,
returning a positive-polarity score $b$.
We construct:
\begin{equation}
    v_0=1-u,\qquad
    v_+=ub,\qquad
    v_-=u(1-b)
\end{equation}
For three-class classification, the resulting score vector is
$\mathbf{r}_t=(v_-,v_0,v_+)$.

For five-class classification, two additional questions assess
strong intensity conditional on positive and negative polarity,
yielding $h_+$ and $h_-$, respectively.
The label scores are
\begin{equation}
\begin{aligned}
    r_{\mathrm{vneg}} &= v_-h_-,\\
    r_{\mathrm{neg}}  &= v_-(1-h_-),\\
    r_{\mathrm{neu}}  &= v_0,\\
    r_{\mathrm{pos}}  &= v_+(1-h_+),\\
    r_{\mathrm{vpos}} &= v_+h_+
\end{aligned}
\label{eq:sentiment_composition}
\end{equation}
This hierarchy separates the existence, direction,
and strength of an evaluation.
The scores are nonnegative and sum to one.

\paragraph{Emotion recognition.}
For each emotion $y_k$, we distinguish evidence for the emotion
from its attribution to the target speaker.
The first question assesses whether the text contains an
expression or appraisal compatible with $y_k$, yielding $a_k$.
The second assesses whether that evidence expresses the
target speaker's state, conditional on its presence, yielding
$b_k$.
We compose the two judgments as $e_k=a_kb_k$.

For multi-label recognition, we set $r_{t,k}=e_k$ without
normalizing across emotions.
For single-label recognition with a neutral category,
an additional question estimates the absence of an
identifiable emotion, yielding $n$.
Writing $\mathcal{E}$ for the nonneutral labels, we define:
\begin{equation}
\begin{aligned}
    r_{t,\mathrm{neutral}} &= n,\\
    r_{t,k}
    &= (1-n)\frac{e_k}{\sum_{j\in\mathcal{E}}e_j},
    \qquad k\in\mathcal{E}
\end{aligned}
\label{eq:emotion_composition}
\end{equation}
If all emotion scores are zero, the nonneutral mass
$1-n$ is distributed uniformly over $\mathcal{E}$.
This construction preserves the neutral score while
allocating the remaining mass among candidate emotions.

\paragraph{Sarcasm detection.}
We assess two cues: a discrepancy between literal wording
and the contextually implied meaning, and rhetorical evidence
supporting an ironic or sarcastic interpretation.
The latter includes contextually inappropriate praise,
exaggeration, and related evaluative devices.
Let $c$ and $a$ denote their respective scores.
We use
\begin{equation}
    r_{t,+}=\frac{c+a}{2},
    \qquad
    r_{t,-}=1-r_{t,+}
\end{equation}

\paragraph{Humor detection.}
We consider four mechanisms: expectation-outcome incongruity,
wordplay, exaggeration or absurdity, and a setup-punchline
relationship.
Each question assesses whether its mechanism supports
a humorous interpretation in context, rather than whether
the mechanism merely occurs.
Given scores $z_1,\ldots,z_4$, we define:
\begin{equation}
    r_{t,+}=\max_{j\in\{1,\ldots,4\}}z_j,
    \qquad
    r_{t,-}=1-r_{t,+}
\end{equation}

\paragraph{Final composition.}
We combine the task-specific scores with direct prediction:
\begin{equation}
    \mathbf{p}^{\mathrm{D}}
    =(1-\lambda)\mathbf{p}^{\mathrm{std}}
      +\lambda\mathbf{r}_t
\label{eq:direct_cue_fusion}
\end{equation}

We fix $\lambda=0.5$ for the main experiments.
Setting $\lambda=0$ recovers Jev-Direct, while $\lambda=1$
uses only the decomposed judgments.

\subsection{Emo-Jev-SC: Multi-Path Judgment Aggregation}
Emo-Jev-SC reduces dependence on a single interpretation
by constructing complementary judgment paths.
Unlike self-consistency prompting \citep{wang2023selfconsistency}, the paths are
predefined sequences of typed probabilistic judgments.
Their diversity comes from the evidence they examine,
instead of repeated sampling of the same question.

\paragraph{Path construction.}
For each task, we define an inventory of ten judgment paths
and select $M$ paths for aggregation.
The main experiments use $M=3$, while the ablation study
evaluates $M\in\{1,3,5,10\}$.

Each perspective defines a fixed sequence
$\mathcal{Q}_t^{(m)}
=(q_{t,1}^{(m)},\ldots,q_{t,L_m}^{(m)})$.
A path maintains a record of previously elicited judgments:
\begin{equation}
\begin{aligned}
    s_{m,0} &= x,\\
    z_{m,\ell}
    &=J_{\theta}(s_{m,\ell-1},q_{t,\ell}^{(m)}),\\
    s_{m,\ell}
    &=\operatorname{Append}
      (s_{m,\ell-1},q_{t,\ell}^{(m)},z_{m,\ell})
\end{aligned}
\label{eq:path_execution}
\end{equation}
where the appended record contains the question semantics and
the returned probabilities.

\paragraph{Path-level prediction.}
After collecting the judgments, each path applies the same
final classification question to the original input and
its accumulated evidence:
\begin{equation}
    \mathbf{p}^{(m)}
    =J_{\theta}(s_{m,L_m},q_t^{\mathrm{final}})
\end{equation}

\paragraph{Probability aggregation.}
We aggregate path-level predictions by equal-weight averaging:
\begin{equation}
    \mathbf{p}^{\mathrm{SC}}
    =\frac{1}{M}\sum_{m=1}^{M}\mathbf{p}^{(m)}
\label{eq:path_aggregation}
\end{equation}

\section{Experiments}
\subsection{Datasets}
We evaluate on eight benchmarks across four tasks:
sentiment analysis with SST-5 \citep{socher2013recursive}
and TweetEval--Sentiment \citep{barbieri2020tweeteval};
emotion recognition with SemEval-2018 Task 1 E-c
\citep{mohammad2018affect} and text-only CMMA
\citep{zhang2023cmma};
sarcasm detection with SemEval-2018 Task 3A
\citep{vanhee2018irony} and English iSarcasmEval Subtask A
\citep{abufarha2022isarcasm};
and humor detection with the Humor Detection (HD) Dataset
\footnote{\url{https://www.kaggle.com/datasets/miadul/humor-detection-dataset}}
and HaHackathon Task 1A \citep{meaney2021humor}.
Table~\ref{tab:datasets} summarizes the dataset statistics.

\begin{table*}[t]
\centering
\small
\setlength{\tabcolsep}{6pt}
\begin{tabular}{llclrr}
\toprule
\textbf{Task} & \textbf{Dataset} & \textbf{Language} & \textbf{Annotation} & \textbf{Labels} & \textbf{Test size} \\
\midrule
Sentiment
& SST-5 & EN & Single-label & 5 & 2,210 \\
& TweetEval--Sentiment & EN & Single-label & 3 & 12,284 \\
\midrule
Emotion
& SemEval-2018 Task 1 E-c & EN & Multi-label & 11 & 3,259 \\
& CMMA (text only) & ZH & Single-label & 7 & 3,961 \\
\midrule
Sarcasm
& SemEval-2018 Task 3A & EN & Binary & 2 & 784 \\
& iSarcasmEval Task A & EN & Binary & 2 & 1,400 \\
\midrule
Humor
& Humor Detection Dataset & EN & Binary & 2 & 50 \\
& HaHackathon Task 1A & EN & Binary & 2 & 1,000 \\
\bottomrule
\end{tabular}
\caption{Dataset statistics for evaluation.}
\label{tab:datasets}
\end{table*}


\subsection{Baselines}
Our experiments compare standard Jev, Emo-Jev, and
five SoTA LLMs: GPT-6 Luna, Gemini 3.8 Flash, Qwen 3.7 Plus, DeepSeek V4.1 Flash,
Doubao-Seed-2.0-mini. Each LLM is evaluated under two prompting conditions.
Input--Output (IO) prompting requests only the predicted label.
Chain-of-Thought (CoT) prompting requests intermediate analysis
followed by the final label. All methods operate in a zero-shot setting without task-specific
fine-tuning or in-context demonstrations. Qwen, DeepSeek, and Doubao use fixed Chinese instructions; GPT-6 Luna
and Gemini use corresponding English instructions. For multi-label prediction, we use a fixed threshold of
$\tau_k=0.5$ for every emotion.


\begin{table*}[t]
\centering
\scriptsize
\setlength{\tabcolsep}{3pt}
\renewcommand{\arraystretch}{1.12}
\resizebox{\textwidth}{!}{%
\begin{tabular}{ll*{16}{c}}
\toprule
\multirow{4}{*}{\textbf{Model}}
& \multirow{4}{*}{\textbf{Setting}}
& \multicolumn{4}{c}{\textbf{Sentiment Analysis}}
& \multicolumn{4}{c}{\textbf{Emotion Recognition}}
& \multicolumn{4}{c}{\textbf{Sarcasm Detection}}
& \multicolumn{4}{c}{\textbf{Humor Detection}} \\
\cmidrule(lr){3-6}
\cmidrule(lr){7-10}
\cmidrule(lr){11-14}
\cmidrule(lr){15-18}
&
& \multicolumn{2}{c}{SST-5}
& \multicolumn{2}{c}{TweetEval}
& \multicolumn{2}{c}{E-c}
& \multicolumn{2}{c}{CMMA}
& \multicolumn{2}{c}{2018 Task 3A}
& \multicolumn{2}{c}{iSarcasm}
& \multicolumn{2}{c}{HD}
& \multicolumn{2}{c}{HaHack} \\
\cmidrule(lr){3-4}
\cmidrule(lr){5-6}
\cmidrule(lr){7-8}
\cmidrule(lr){9-10}
\cmidrule(lr){11-12}
\cmidrule(lr){13-14}
\cmidrule(lr){15-16}
\cmidrule(lr){17-18}
&
& Acc & Ma-F1
& Acc & Ma-F1
& Acc & Ma-F1
& Acc & Ma-F1
& Acc & Ma-F1
& Acc & Ma-F1
& Acc & Ma-F1
& Acc & Ma-F1 \\
\midrule
\multirow{2}{*}{Qwen 3.7 Plus}
& IO
& \underline{59.55} & \underline{55.57}
& 67.98 & 68.11
& 15.65 & 47.51
& 47.74 & 21.47
& \textbf{87.12} & \textbf{86.99}
& 78.21 & 70.29
& \textbf{88.00} & \textbf{87.92}
& 93.70 & 93.38 \\
& CoT
& 56.79 & 53.23
& 66.41 & 66.62
& 15.50 & 48.72
& 54.53 & 21.65
& \underline{86.99} & \underline{86.85}
& \underline{84.14} & \underline{76.04}
& 78.00 & 77.26
& 94.00 & 93.69 \\
\addlinespace
\multirow{2}{*}{DeepSeek V4.1 Flash}
& IO
& 56.83 & 53.85
& \underline{70.42} & 70.39
& 14.51 & \textbf{52.61}
& 55.97 & \textbf{23.31}
& 83.42 & 83.33
& 71.64 & 64.64
& 76.00 & 74.53
& 94.60 & 94.21 \\
& CoT
& 55.25 & 50.52
& 67.66 & 67.81
& 14.48 & 51.68
& 54.35 & 21.46
& 75.89 & 77.91
& 64.21 & 59.83
& 66.00 & 64.27
& 93.00 & 92.69 \\
\addlinespace
\multirow{2}{*}{Doubao-Seed-2.0-mini}
& IO
& 52.04 & 46.41
& \textbf{71.85} & \textbf{71.16}
& 14.73 & 44.10
& \textbf{61.88} & 21.01
& 84.31 & 83.01
& 80.64 & 70.19
& 60.00 & 59.42
& 73.50 & 73.50 \\
& CoT
& 48.24 & 42.07
& 70.37 & 69.88
& 13.04 & 44.26
& \underline{60.59} & 19.82
& 85.84 & 85.64
& 81.00 & 71.00
& 74.00 & 73.91
& 71.80 & 72.02 \\
\addlinespace
\multirow{2}{*}{GPT-6 Luna}
& IO
& 55.48 & 50.07
& 70.38 & \underline{70.48}
& 15.68 & 50.73
& 50.82 & 21.98
& 78.95 & 79.00
& 72.79 & 65.34
& 62.00 & 55.59
& 94.20 & 93.84 \\
& CoT
& 56.11 & 51.02
& 70.29 & 70.27
& 14.94 & 50.55
& 50.62 & 21.46
& 79.34 & 79.38
& 73.29 & 65.78
& 66.00 & 61.56
& 94.90 & 94.63 \\
\addlinespace
\multirow{2}{*}{Gemini 3.8 Flash}
& IO
& 59.32 & 54.91
& 68.72 & 68.87
& \textbf{19.15} & \underline{52.52}
& 47.87 & \underline{22.56}
& 85.71 & 85.61
& 83.29 & 75.42
& \underline{82.00} & \underline{81.99}
& \underline{96.60} & \underline{96.38} \\
& CoT
& \textbf{60.09} & \textbf{55.59}
& 68.56 & 68.73
& \underline{18.47} & 52.20
& 48.17 & 22.33
& 86.48 & 86.32
& \textbf{84.79} & \textbf{77.10}
& 74.00 & 73.99
& \textbf{96.80} & \textbf{96.60} \\
\midrule
Jev-Direct
& Direct
& 56.24 & 52.40
& 68.30 & 68.51
& 11.41 & 52.47
& 52.28 & 21.06
& 83.67 & 83.61
& 79.57 & 70.71
& 64.00 & 61.80
& 93.30 & 92.90 \\
\textbf{Emo-Jev-D}
& Decomposed
& 55.25 & 51.87
& 67.43 & 67.60
& 10.71 & 52.15
& 54.86 & 21.53
& 81.63 & 81.61
& 76.71 & 68.08
& 64.00 & 60.94
& 92.70 & 92.21 \\
\textbf{Emo-Jev-SC}
& SC
& 57.42 & 53.55
& 68.51 & 68.60
& 12.18 & 42.00
& 51.81 & 20.84
& 83.93 & 83.85
& 76.64 & 68.15
& 66.00 & 63.53
& 92.80 & 92.34 \\
\bottomrule
\end{tabular}%
}
\caption{Classification performance (\%) on eight datasets
grouped by task. The best and second-best results in each metric
column are marked in bold and underlined, respectively.
Jev uses \texttt{jev-1.13.0}; D uses $\lambda=0.5$,
and SC uses $M=3$ with path-selection seed 42.  }
\label{tab:main_results}
\end{table*}

\subsection{Main Results}
\label{sec:main_results}

Table~\ref{tab:main_results} reports classification performance
on eight benchmarks. 

\paragraph{IO versus CoT prompting.}
CoT improves macro-F1 in 17 of the 40 model--dataset
comparisons and reduces it in the remaining 23.
Across all five models and eight datasets, average
macro-F1 decreases from 63.56\% under IO to 62.91\%
under CoT, a relative reduction of 1.02\%.
Doubao and GPT-6 Luna benefit on average, where
Qwen, DeepSeek, and Gemini perform worse.
DeepSeek declines on every dataset, with average
macro-F1 falling from 64.61\% to 60.77\%.
Thus, CoT does not consistently improve classification
under the evaluated settings.

\paragraph{Standard Jev versus LLMs.}
Standard Jev achieves 62.93\% average macro-F1,
exceeding five of the ten generative model--prompting
configurations.
However, it trails the strongest configuration,
Gemini under IO (67.28\%), by 6.47\% in relative terms
and does not rank first on any dataset.
Its performance is closest to the best generative
result on E-c (52.47\% versus 52.61\%) but falls
substantially behind on Humor (61.80\% versus 87.92\%).
Standard Jev is therefore competitive on some benchmarks
without consistently matching the strongest generative
models.

\paragraph{Comparison within the Jev family.}
Emo-Jev-D achieves 62.00\% average macro-F1,
a relative decrease of 1.48\% from Standard Jev.
It improves only on CMMA, where macro-F1 increases
from 21.06\% to 21.53\% and accuracy from 52.28\%
to 54.86\%.
Macro-F1 declines on the other seven datasets,
with the largest decrease on iSarcasm
(70.71\% to 68.08\%).
The current decomposition and composition rules
therefore provide no consistent improvement.

Emo-Jev-SC achieves 61.61\% average macro-F1,
a relative decrease of 2.11\% from Standard Jev.
It improves on SST-5, TweetEval, Sarcasm, and HD,
but its E-c macro-F1 drops from 52.47\% to 42.00\%.
Excluding E-c, average macro-F1 is nearly unchanged:
64.43\% for Standard Jev and 64.41\% for SC.
The overall decline is thus concentrated in E-c,
while improvements elsewhere are offset by losses
on other datasets.

\subsection{Ablation Study}
\label{sec:ablation}
We examine probability composition in Emo-Jev-D
and the effect of path count in Emo-Jev-SC.

\paragraph{Emo-Jev-D.}
We compare three configurations: Standard Jev
($\lambda=0$), decomposition only ($\lambda=1$),
and their equal-weight combination ($\lambda=0.5$).
This comparison assesses the effectiveness of
decomposed judgments both independently and
in combination with standard predictions.

As shown in Table~\ref{tab:ablation_d}, decomposition
alone reduces average macro-F1 from 62.93\% to 57.31\%,
a relative decrease of 8.93\%, with declines on seven
of the eight datasets.
Combining both predictions improves over decomposition
alone on every dataset, raising average macro-F1
to 62.00\%, a relative increase of 8.18\%.
Nevertheless, the combined method remains below
Standard Jev on seven datasets and improves only
on CMMA (21.06\% to 21.53\%).
Fusion therefore mitigates the losses from decomposition
without consistently outperforming standard prediction.

\begin{table}[t]
\centering
\small
\setlength{\tabcolsep}{5pt}
\begin{tabular}{lccc}
\toprule
Dataset
& Standard
& Decomp.\ only
& Full D \\
& $\lambda=0$
& $\lambda=1$
& $\lambda=0.5$ \\
\midrule
SST-5 & 52.40 & 45.84 & 51.87 \\
TweetEval & 68.51 & 65.01 & 67.60 \\
E-c & 52.47 & 50.47 & 52.15 \\
CMMA & 21.06 & 21.13 & 21.53 \\
Irony & 83.61 & 76.01 & 81.61 \\
iSarcasm & 70.71 & 63.28 & 68.08 \\
Humor & 61.80 & 47.62 & 60.94 \\
HaHack & 92.90 & 89.14 & 92.21 \\
\midrule
Average & 62.93 & 57.31 & 62.00 \\
\bottomrule
\end{tabular}
\caption{Ablation of Emo-Jev-D.
Values are macro-F1 (\%).}
\label{tab:ablation_d}
\end{table}

\begin{table}[t]
\centering
\small
\setlength{\tabcolsep}{5pt}
\begin{tabular}{lcccc}
\toprule
Dataset & $M=1$ & $M=3$ & $M=5$ & $M=10$ \\
\midrule
SST-5     & 52.94 & 52.73 & 52.74 & 53.03 \\
TweetEval & 67.02 & 67.14 & 67.13 & 67.43 \\
E-c       & 42.63 & 42.16 & 43.26 & 42.82 \\
CMMA      & 20.33 & 21.03 & 21.00 & 21.09 \\
Irony     & 82.79 & 85.17 & 84.90 & 85.23 \\
iSarcasm  & 69.72 & 69.67 & 70.45 & 70.10 \\
Humor     & 60.40 & 61.96 & 58.78 & 63.53 \\
HaHack    & 91.13 & 92.33 & 92.30 & 92.35 \\
\midrule
Average   & 60.87 & 61.52 & 61.32 & 61.95 \\
\bottomrule
\end{tabular}
\caption{Effect of the number of judgment paths in
Emo-Jev-SC.}
\label{tab:ablation_sc}
\end{table}

\paragraph{Emo-Jev-SC.}
We vary the number of judgment paths
$M\in\{1,3,5,10\}$ while keeping the inference procedure
and aggregation rule fixed.
Paths are selected from a predefined ten-path inventory
using nested subsets.
We report mean macro-F1 across five path-selection
seeds (42--46), reusing the same cached path outputs.
At $M=10$, every seed uses the complete inventory.

As shown in Table~\ref{tab:ablation_sc}, average macro-F1
exhibits a gradual overall increase as more judgment
paths are aggregated, rising from 60.87\% at $M=1$
to 61.95\% at $M=10$, a relative improvement of 1.77\%.
Although a small decrease occurs between $M=3$ and
$M=5$, the full inventory yields the highest average
score and the best performance on six of eight datasets.
The modest overall gain suggests that additional
judgment paths can improve classification.


\begin{figure*}[t]
    \centering
    \includegraphics[width=0.95\linewidth]{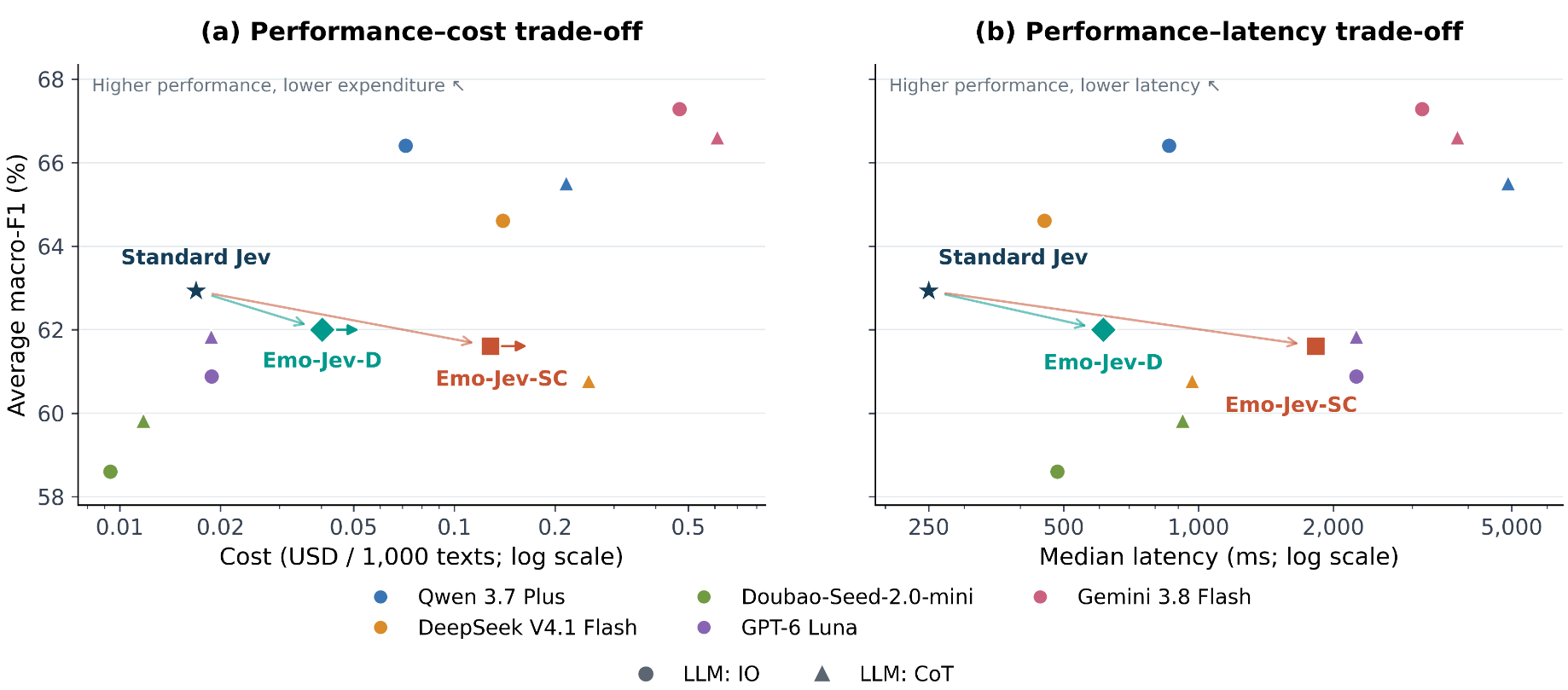}
    \caption{Performance--efficiency trade-offs across eight benchmarks:
(a) average macro-F1 versus cost per 1,000 texts;
(b) average macro-F1 versus measured median latency.
Both horizontal axes use logarithmic scales.
Positions closer to the upper-left indicate better trade-offs.}
    \label{fig:efficiency}
\end{figure*}

\subsection{Efficiency Analysis}
\label{sec:efficiency}

Table~\ref{tab:efficiency} and Fig.~\ref{fig:efficiency} compares token usage, latency,
and monetary cost across eight benchmarks.

\paragraph{Token usage.}
Standard Jev does not consistently save tokens.
It records 403.02 input and 64.53 output tokens per
instance, exceeding both counts of Qwen, DeepSeek,
and Doubao under IO.
Unlike a label-only response, its structured output
includes field names and probability values,
which contribute to the reported token count.
CoT increases output length for four of the five
generative models, most notably Qwen, whose count
rises from 5.68 to 327.88 tokens.
Within the Jev family, D and SC increase input usage
to 960.63 and 3,052.91 tokens and output usage to
186.56 and 362.37 tokens, respectively.
The additional questions, intermediate probability
records, and repeated inputs across paths account
for this overhead.

\paragraph{Latency.}
Standard Jev has the lowest reported median and P95
latency, at 250 and 313 ms, respectively.
These values are 44.81\% and 55.48\% lower than
DeepSeek IO, the fastest generative configuration
in the table.
This observation is consistent with Jev's direct
decision interface, which avoids generating extended
textual responses.
CoT increases median latency for four generative
models, with Qwen rising from 859 to 4,906 ms
alongside its substantial increase in output length.
D and SC raise Jev's reported median latency to
613 and 1,826 ms, respectively.
The additional judgments increase inference work,
while SC also introduces dependent stages within
each path that cannot all be executed concurrently.

\paragraph{Monetary cost.}
Standard Jev costs \$0.0169 per
1,000 texts, less than every evaluated generative
configuration except Doubao-Seed-2.0-mini.
Its cost is 76.38\% lower than Qwen IO and
96.41\% lower than Gemini IO.
The savings arise from the applied pricing,
including free outputs, rather than consistently
lower token usage.
CoT increases cost for four generative models,
reflecting the additional billed usage.
For Jev, the extra input required by D and SC
raises costs to at least \$0.0403
and \$0.1282, approximately 2.38 and 7.59 times
the standard cost.
These are lower bounds because some retry usage
is missing.
Overall, Standard Jev offers low observed latency
and competitive cost without consistently saving
tokens.
    
    \begin{table*}[t]
    \centering
    \small
    \setlength{\tabcolsep}{6pt}
    \begin{tabular}{llrrrrr}
    \toprule
    Model & Setting
    & Input tokens & Output tokens
    & Median (ms) & P95 (ms) & Cost / 1K(\$) \\
    \midrule
    Qwen 3.7 Plus
    & IO & 156.07 & 5.68 & 859 & 1,187 & 0.07155 \\
    & CoT & 159.87 & 327.88 & 4,906 & 17,088 & 0.21577 \\
    DeepSeek V4.1 Flash
    & IO & 143.45 & 7.82 & 453 & 703 & 0.13958 \\
    & CoT & 147.45 & 125.60 & 968 & 2,094 & 0.25196 \\
    Doubao-Seed-2.0-mini
    & IO & 177.29 & 6.54 & 484 & 1,172 & 0.00937 \\
    & CoT & 182.29 & 48.71 & 922 & 1,656 & 0.01177 \\
    GPT-6 Luna
    & IO & 2,018.68 & 10.69 & 2,250 & 20,573 & 0.01883 \\
    & CoT & 2,012.61 & 10.69 & 2,250 & 17,619 & 0.01877 \\
    Gemini 3.8 Flash
    & IO & 393.70 & 194.90 & 3,156 & 9,250 & 0.47043 \\
    & CoT & 398.70 & 364.56 & 3,781 & 10,079 & 0.61002 \\
    \midrule
    Jev-Direct
    & Direct & 403.02 & 64.53 & 250 & 313 & 0.0169 \\
    Emo-Jev-D
    & Decomposed & 960.63 & 186.56 & 613 & 768 & $\geq 0.0403$ \\
    Emo-Jev-SC
    & SC & 3,052.91 & 362.37 & 1,826 & 2,286 & $\geq 0.1282$ \\
    \bottomrule
    \end{tabular}
    \caption{Inference efficiency across eight benchmarks in terms of token usage, latency, and monetary cost. }
    \label{tab:efficiency}
    \end{table*}

\section{Conclusion}
We evaluated Jev on eight datasets spanning sentiment,
emotion, sarcasm, and humor classification, and proposed
Emo-Jev-D and Emo-Jev-SC for probabilistic composition
and multi-path aggregation.
Standard Jev achieved competitive performance on
several benchmarks, with low observed latency and
generally lower monetary cost than the evaluated LLMs.
However, it did not consistently save tokens or match
the strongest baseline's classification performance.
Neither Emo-Jev variant improved average macro-F1,
despite gains on individual datasets.
These results highlight Jev's practical efficiency
while showing that additional probabilistic judgments
do not necessarily improve prediction quality.

\section*{Limitations}
Our evaluation covers two datasets per task family
and uses only textual inputs.
The findings may not generalize to other domains,
languages, or multimodal settings.
Additionally, Emo-Jev uses predefined question
inventories and fixed composition rules.
Alternative decompositions and aggregation strategies
may yield different performance--cost trade-offs.

\bibliography{references}

\end{document}